\documentclass{article}

\PassOptionsToPackage{numbers, compress}{natbib}

\usepackage[normalem]{ulem}
\useunder{\uline}{\ul}{}
\usepackage[T1]{fontenc}
\usepackage{bbding}
\usepackage{amsmath}

\usepackage{graphicx}

\usepackage[preprint]{neurips_2024}

\usepackage[utf8]{inputenc} 
\usepackage[T1]{fontenc}    
\usepackage{hyperref}       
\usepackage{url}            
\usepackage{booktabs}       
\usepackage{amsfonts}       
\usepackage{nicefrac}       
\usepackage{microtype}      
\usepackage{xcolor}         

\title{ProMISe: Promptable Medical Image Segmentation using SAM}

\author{Jinfeng Wang$^{1,2,4\thanks{These authors contributed equally to this work.}}$, Sifan Song$^{3\footnotemark[1]}$, Xinkun Wang$^{1\footnotemark[1]}$, Yiyi Wang$^{1\footnotemark[1]}$, Yiyi Miao$^{1,2}$, \And Jionglong Su$^{1\footnotemark[2]\thanks{Corresponding authors: Jionglong Su ({\tt\small Jionglong.Su@xjtlu.edu.cn}), S. Kevin Zhou ({\tt\small skevinzhou@ ustc.edu.cn})}}$, S. Kevin Zhou$^{4,5\footnotemark[2]}$ \and \\
$^1$ \small School of AI and Advanced Computing, XJTLU Entreprenur College (Taicang), \\ 
     \small Xi’an Jiaotong-liverpool University, Suzhou, China\\
$^2$ \small University of Liverpool, Liverpool, UK\\
$^3$ \small Center for Advanced Medical Computing and Analysis (CAMCA), \\ 
     \small Massachusetts General Hospital and Harvard Medical School, Boston, USA\\
$^4$ \small School of BME \& Suzhou Institute for Advanced Research,\\
     \small Center for Medical Imaging, Robotics, Analytic Computing \& Learning (MIRACLE),\\
     \small University of Science and Technology of China, Suzhou, China\\
$^5$ \small Key Lab of Intelligent Information Processing of Chinese Academy of Sciences (CAS),\\
     \small Institute of Computing Technology, CAS, Beijing, China
     }

\begin{document}

\maketitle

\begin{abstract}
With the proposal of Segment Anything Model (SAM), fine-tuning SAM for medical image segmentation (MIS) has become popular. However, due to the large size of the SAM model and the significant domain gap between natural and medical images, fine-tuning-based strategies are costly with potential risk of instability, feature damage and catastrophic forgetting. Furthermore, some methods of transferring SAM to a domain-specific MIS through fine-tuning strategies disable the model's prompting capability, severely limiting its utilization scenarios. In this paper, we propose an Auto-Prompting Module (APM), which provides SAM-based foundation model with Euclidean adaptive prompts in the target domain. Our experiments demonstrate that such adaptive prompts significantly improve SAM's non-fine-tuned performance in MIS. In addition, we propose a novel non-invasive method called Incremental Pattern Shifting (IPS) to adapt SAM to specific medical domains. Experimental results show that the IPS enables SAM to achieve state-of-the-art or competitive performance in MIS without the need for fine-tuning. By coupling these two methods, we propose \textbf{ProMISe}, an end-to-end non-fine-tuned framework for \textbf{Pro}mptable \textbf{M}edical \textbf{I}mage \textbf{Se}gmentation. Our experiments demonstrate that both using our methods individually or in combination achieves satisfactory performance in low-cost pattern shifting, with all of SAM's parameters frozen. \textit{Code is available at \href{https://github.com/xinkunwang111/ProMISe}{https://github.com/xinkunwang111/ProMISe}}
\end{abstract}

\section{Introduction}
Recently, many large-scale foundation models \cite{kirillov2023segment,Wang_Zhang_Cao_Wang_Shen_Huang} deliver exciting performance with promising results in various domains. Among them, the latest one SAM \cite{kirillov2023segment}, a foundation model for natural image segmentation, has attracted significant attention from researchers in the computer vision community. SAM enables interactive segmentation of targets through prompts such as points, bounding boxes, masks, and text. Medical Image Segmentation (MIS) tasks are crucial in medical image analysis. Unlike natural images, medical images contain diverse and distinctive data modalities and application scenarios. The transfer and deployment of SAM for MIS present a meaningful and promising research area.

Some studies \cite{ccheng2023sam,zhou2023can,mazurowski2023segment,he2023accuracy} have explored the capability of SAM in MIS in a zero-shot fashion. Unfortunately, their results demonstrate that neither points or bounding boxes guided by ground truth (GT), nor the automated methods of SAM, can achieve  satisfactory and practical performance in MIS. Also, different studies have used various prompts (such as latent representations \cite{shaharabany2023autosam,zhang2023customized}, bounding boxes \cite{li2023polyp,cheng2023sam,wu2023medical}, text \cite{wu2023medical,paranjape2023adaptivesam}, mask \cite{cheng2023sam}, points \cite{cheng2023sam,wu2023medical}) by different strategies (\textit{e.g.}, all-parameter fine-tuning \cite{li2023polyp}, adapter \cite{wu2023medical,chen2023sam}, bias-tuning \cite{paranjape2023adaptivesam}, and LoRA \cite{zhang2023customized}) to different degrees (decoder-only \cite{ma2024segment}, prompt-encoder-only \cite{shaharabany2023autosam}, and all components \cite{li2023polyp}), with the aim of fine-tuning SAM for transfer to the target domain. However, since SAM is a large foundation model trained on extensive datasets, these fine-tuning-based approaches incur enormous training costs and face potential risks of instability, feature damage and catastrophic forgetting. Such fine-tuning-based transfer methods typically utilize the model as a heavy pre-trained model rather than a foundation model.

\begin{figure}[t]
\includegraphics[width=\textwidth]{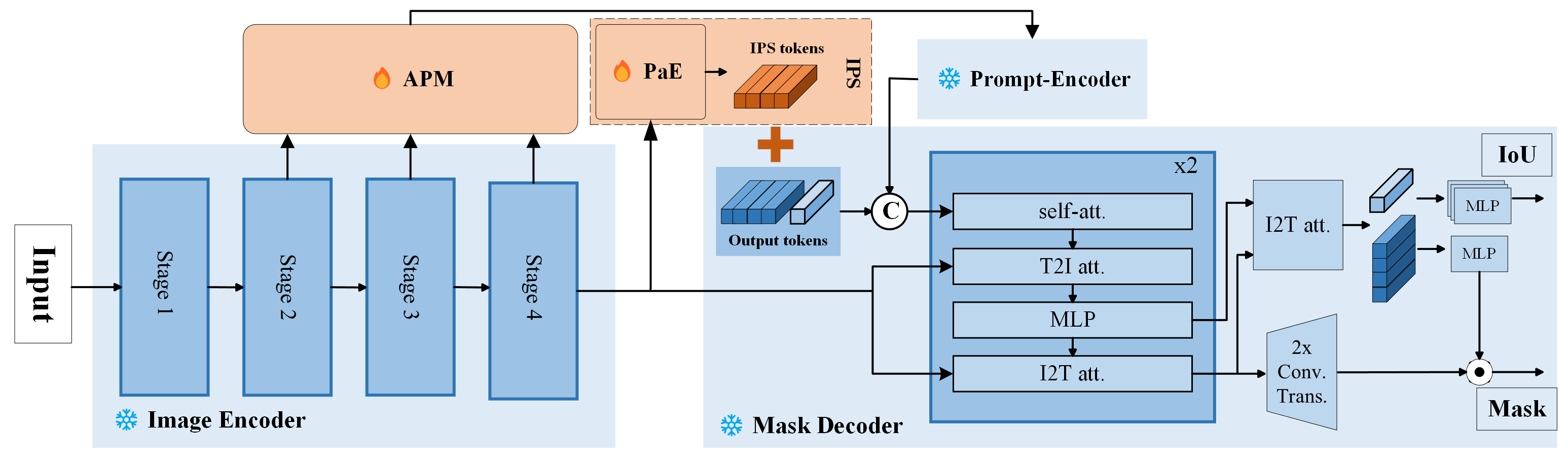}
\caption{Overview of ProMISe. All three components of the original SAM are frozen. The \textbf{A}uto-\textbf{P}rompting \textbf{M}odule (APM) leverages features from the image encoder to predict optimal prompts in Euclidean space. The \textbf{Pa}ttern \textbf{E}mbedding (PaE) module analyzes the image embedding to extract pattern gaps between the target and source domains. The \textbf{I}ncremental \textbf{P}attern \textbf{S}hifting (IPS) tokens are added to the mask tokens of the output tokens to realize the decoder's shifting of mask patterns.} \label{Overview}
\end{figure}

\begin{figure}[t]
\includegraphics[width=\textwidth]{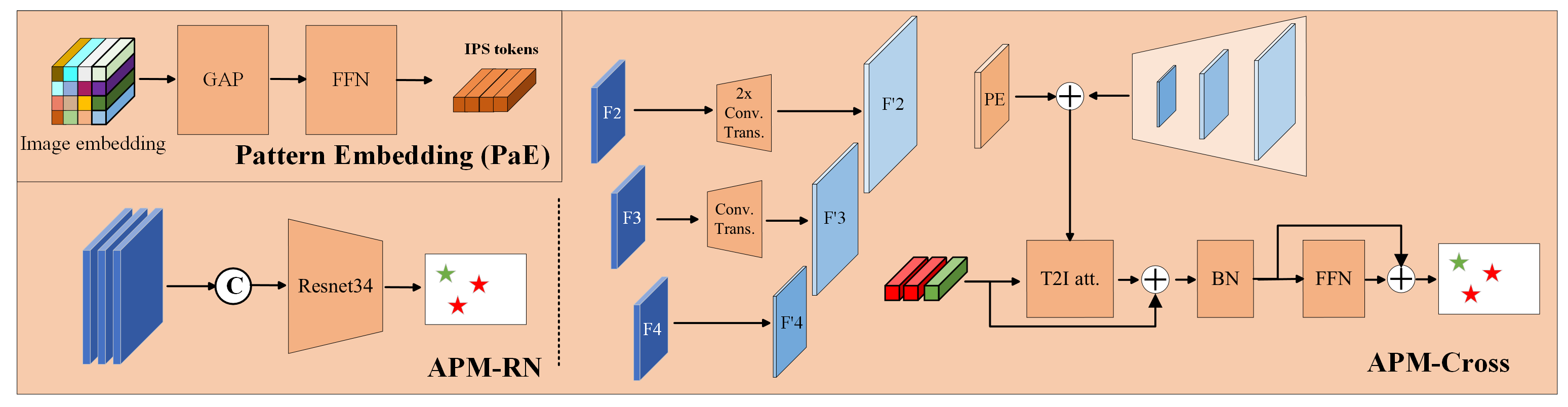}
\caption{Detailed structure of APM and PaE. The APM can be implemented using various operators and modules, such as CNN and Transformer.} \label{Overview2}
\end{figure}

In this paper, we rethink the use of each component of SAM in the following way. First, the prompt encoder, as a mapping between Euclidean and latent space, has been adequately trained during the training process of SAM. Some studies attempt to use latent representations instead of explicit Euclidean prompts to automate the SAM model, but this approach sacrifices its interactive capability and interpretability. Second,  as a foundation model trained on extensive datasets with a complex training strategy, the encoder of SAM already possesses robust feature extraction capabilities. We argue that utilizing these capabilities efficiently and properly in unseen domains is more valuable than fine-tuning them. Third, by analyzing the mask decoder, we observe that the output tokens provide prior pattern knowledge for mask prediction based on prompts. Therefore, we claim that pattern shifting of output tokens can achieve more efficient and stable domain adaptation for SAM compared to fine-tuning.

As demonstrated in Fig. \ref{Overview} and Fig. \ref{Overview2}, based on the above arguments: 1) We propose the {A}uto-{P}rompting {M}odule (APM), which leverages the SAM framework for training and provides adaptive Euclidean prompts for SAM; 2) We propose {I}ncremental {P}attern {S}hifting (IPS), a novel non-invasive pattern shifting method which couples a {Pa}ttern {E}mbedding (PaE) module with IPS tokens to cost-effectively shift the prior pattern knowledge of the mask decoder. The IPS method enables SAM to achieve state-of-the-art (SOTA) and competitive performance without the need for fine-tuning; 3) We couple the above two methods to propose the \textbf{Pro}mptable \textbf{M}edical \textbf{I}mage \textbf{Se}gmentation (\textbf{ProMISe}) framework utilizing SAM. Notably, this method is able to transfer SAM to MIS while keeping \textbf{all of SAM's parameters frozen}, resulting in significantly reduced training costs, improved stability, and the ability to retain spatial prompting capability.

\section{Adaptive Prompt} \label{Sec: APM}

\subsection{Motivation}
Several zero-shot studies \cite{ccheng2023sam,zhou2023can,mazurowski2023segment,he2023accuracy} utilize the interactive capability of SAM to investigate its potential as a large-scale foundation model for transferring to MIS tasks in an untrained manner. However, most of these approaches employ GT-based prompts, which are generated from GT masks using various different methods, such as GT-based foreground-background point sampling and noisy GT-bounding box. However, these GT-based prompt strategies fail to enable SAM to achieve practical performance in MIS tasks. We argue that this limitation is due to the coupling of prompts, image embeddings and mask patterns. In other words, without effectively utilizing image embeddings and transferring mask patterns, it remains challenging for SAM to consistently deliver satisfactory performance in unseen domains by only relying on GT-based spatial prompts.

\subsection{Proposed Method}
As shown in Fig. \ref{Overview} and \ref{Overview2}, we design a lightweight end-to-end module, called Auto-Prompting Module (APM), which effectively integrates the multi-level features of SAM's image encoder to predict optimal prompts in Euclidean space. By generating adaptive prompts that provide more fine-grained spatial information, the APM significantly enhances the untrained performance of SAM in the target domain. This indicates the highly expressive feature extraction capability of SAM, even in previously unseen domains.

During our research, we notice a significant performance limitation when using bounding boxes as prompts. We attribute this to the fact that bounding boxes only provide coarse-grained location information and lack fine-grained details. Thus, we mainly utilize point-based prompts in this paper. It is worth noting that our method is applicable to any form of prompts except text, and the APM can be implemented using various operators and modules (\textit{e.g.}, ResNet34 and a simple cross-attention block).

\section{Incremental Pattern Shifting}

\subsection{Rethink the mask decoder}
As mentioned above, many studies based on fine-tuning strategies have been proposed to improve the performance of SAM in MIS tasks. However, inappropriate fine-tuning is likely to weaken the original parameter distribution of SAM. As such, it becomes difficult to sufficiently adjust to the optimal target distribution, especially in medical datasets with distinctive domains, making fine-tuning SAM unstable and inefficient. From a broader perspective, a simple fine-tuning strategy for pattern shifting is regarded as the model as a heavily pre-trained model rather than a foundation model, resulting in degraded model capabilities.

\begin{figure}[!ht]
\centering
\includegraphics[width=\columnwidth]{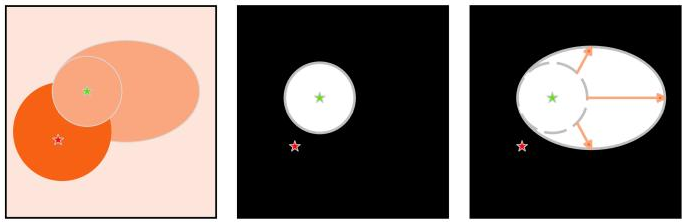}
\caption{Theoretical illustration of IPS. Left: Image with point prompts; Middle: Output mask from vanilla SAM; Right: Output mask from SAM with IPS. Arrows represent patterns shifting.} 
\label{fig: IPS}
\end{figure}

After rethinking the components of the SAM, we observe that the SAM has a very efficient mask decoder that refers to DETR \cite{carion2020end} and MaskFormer \cite{cheng2021per}. In this study, we find that in the mask decoder, output tokens initially receive the location information of the point of interest (PoI) or region of interest (RoI) from the prompt encoder through self-attention. Subsequently, as shown in Fig. \ref{Overview}, output tokens extract semantic information from the image embedding through cross-attention, forming semantic patterns. 

\subsection{Proposed Method}
Based on these findings, we argue that with a large amount of training, the output tokens can be regarded as the \textit{pattern information} acquired from the source domain, enabling SAM to generate predicted masks based on PoI or RoI. Therefore, assuming the image encoder of SAM is sufficiently powerful and well-trained, transferring SAM to MIS can be equivalent to transferring the mask prediction patterns of SAM to MIS.

\begin{gather}
    [IoU, Mask] = \text{MaskDecoder}({ImageEmbedding}, T_{Pattern}, T_{Prompts})  \label{Equ: MaskD}
\end{gather}

\begin{gather}
    T_{Pattern} = \text{Concat}(T_{IoU}, (T_{Mask}+ T_{IPS})) \label{Equ: PatternT}
\end{gather}

\begin{gather}
    T_{IPS} = \text{FFN}(\text{GAP}(ImageEmbedding)) \label{Equ: IPST}
\end{gather}

With the above arguments, we propose a method called Incremental Pattern Shifting (IPS). As shown in Fig. \ref{Overview}, we extract the \textit{pattern shifting information} (IPS tokens) from the image embedding using a lightweight PaE module, and then non-invasively shift the patterns of the mask decoder by adding the IPS tokens to the mask tokens (Eqs. \ref{Equ: MaskD}-\ref{Equ: IPST}). In Fig. \ref{fig: IPS}, the mask decoder, when provided with the same PoI, generates mask predictions that are more compatible with the target domain after pattern shifting. As an experimental validation of Fig. \ref{fig: IPS}, the right two columns of Fig. \ref{fig:IPS-points} (SAM and IPS) show that the transfer of the segmentation pattern of SAM using only IPS is solid. Notably, our experiments show the effectiveness of IPS, utilizing GT-based prompts for both training and inference, in enabling SAM to achieve SOTA or competitive performance in MIS tasks.

\section{ProMISe framework} \label{sam-rider}

As shown in Fig. \ref{Overview}, IPS does not conflict with the proposed APM which provides prompts for end-to-end SAM transfer and inference. Deploying non-invasive APM, PaE, and IPS tokens to SAM allows automation and transfer of original SAM to MIS tasks. Our proposed IPS method effectively facilitates segmentation pattern shifting. However, using GT-based prompts for pattern transfer does not allow end-to-end training process. In addition, GT-based prompt sampling may introduce randomness, thereby posing challenges to the stability of the training and evaluation process. Therefore, we couple the IPS method with APM to achieve end-to-end pattern shifting for SAM to target medical domains. Moreover, favored by the preservation of Euclidean spatial form prompts, this ProMISe framework can handle both automatic and manual prompts during inference while retaining its interpretability.


\section{Experiments}
\subsection{Experimental Setup}
We individually conduct experiments in two modalities: endoscopy and dermoscopy. For endoscopy experiments, we follow the experimental setups of PraNet \cite{fan2020pranet}, DuAT \cite{tang2023duat},  and SSFormer \cite{wang2022stepwise}. We extract 1450 images from Kavsir \cite{Jha_Smedsrud_Riegler_Halvorsen_Lange_Johansen_Johansen_2019} and CVC-ClinicDB \cite{Silva_Histace_Romain_Dray_Granado_2014} as the training set. Tests are then conducted on Kvasir, EndoScene \cite{tajbakhsh2015automated}, CVC-ColonDB \cite{Bernal_Sánchez_Fernández-Esparrach_Gil_Rodríguez_Vilariño_2015}, and ETIS \cite{vazquez2017benchmark}, using Mean Absolute Error (MAE), mean Dice, and mean IoU as metrics. For dermatoscopy, we use ISIC2018 for training and testing, with mean Dice and mean IoU as metrics. We utilize ViT-B as the image encoder for all SAM-related methods in the experiments. In addition, we apply three prompt settings in training and inference, namely 3/5/16 points (1/2/8  positive + 2/3/8 negative, 3/5/16P, respectively).  All GT-based prompt points used in this paper were obtained by random sampling from GT masks. For fair comparisons, the prompt points provided to the models are exactly the same in each experimental setting.

We implement our methods in pytorch, using one TESLA A100 (80G) GPU. Adam optimizer and a learning rate of 0.00001 are employed. The training period is 200 epochs. Our loss is a combination of Dice and BCE loss. The IPS method requires approximately 22 GPU hours and 25 GPU hours for training in polyps and ISIC, respectively. In addition, APM-cross requires approximately 28 GPU hours and 30 GPU hours for training in polyps and ISIC. The training periods for APM-RN are 21 GPU hours and 25 GPU hours.

\begin{figure}[t]
    \centering
    \includegraphics[width=\textwidth]{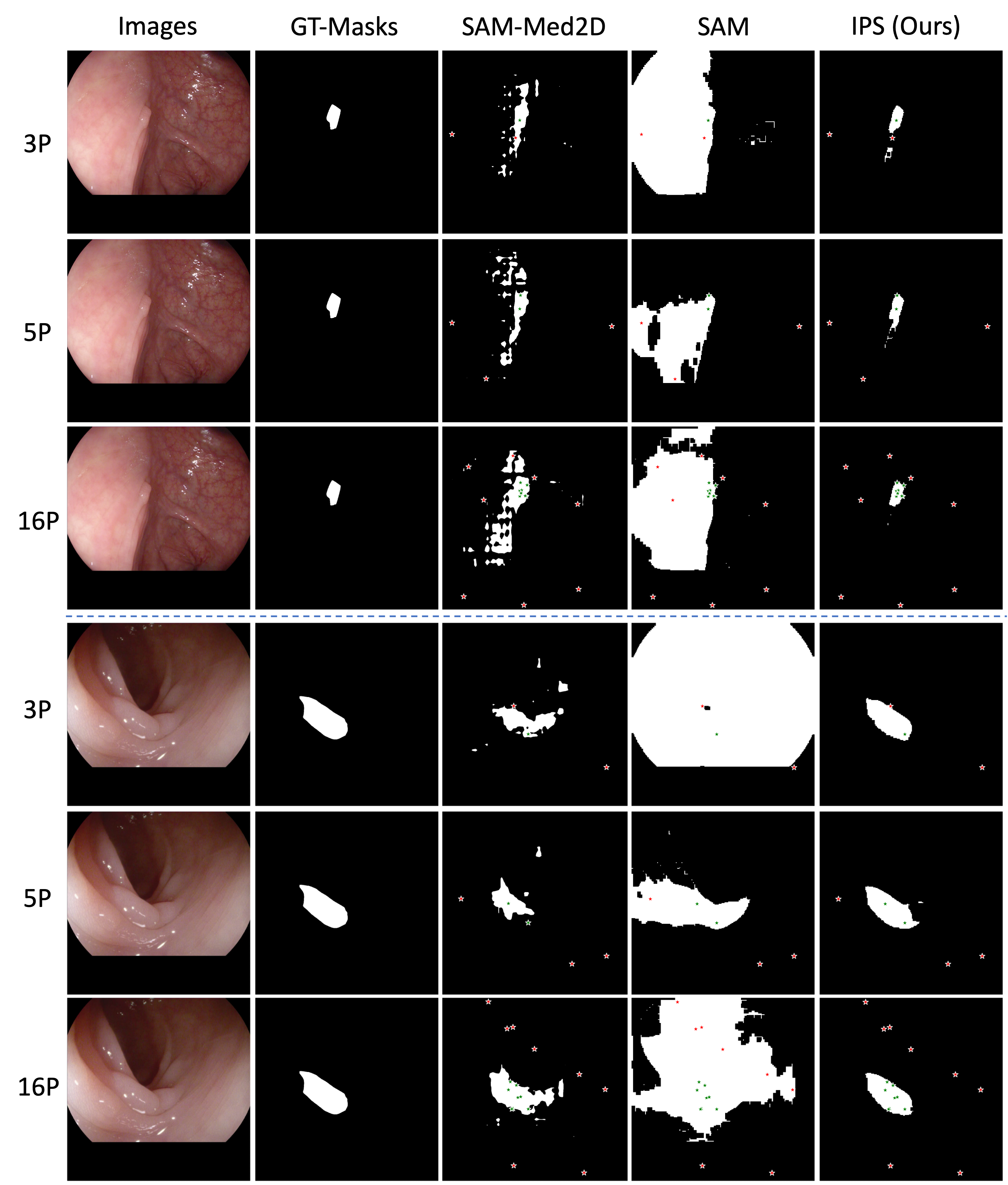}
    \caption{Comparison of the performance with different prompt number in endoscopy datasets. The 3P, 5P and 16P contain 1, 2, and 8 positive, as well as 2, 3, and 8 negative points, respectively.}
    \label{fig:IPS-points}
\end{figure}

\subsection{Adaptive Prompt}\label{subsection:APM}
As described in Sect. \ref{Sec: APM}, to study the actual performance of vanilla SAM in MIS, we train and test two types of APMs, \textit{i.e.}, ResNet34 (RN) and a one-layer cross-attention Transformer module (Cross), on polyp benchmarks. As shown in Tab. \ref{APM}, the adaptive prompts generated by both APMs effectively improve the performance of SAM in the polyp segmentation task, achieving a level comparable to the baseline MIS segmentation model. Furthermore, we observe a significant reduction in MAE by increasing the number of point prompts with more provided fine-grained information, which suggests that our method effectively utilizes the boundary sensitivity of vanilla SAM.

\begin{table}[!t]
\centering
\caption{Comparison of the performance enhancement brought to SAM by adaptive prompts provided by APMs (RN and Cross). $^{a}$ Trainable Parameters.} 
\label{APM}
\resizebox{\columnwidth}{!}{
\begin{tabular}{l|c|ccc|ccc|ccc}
\hline
\multicolumn{2}{c|}{Benchmarks}         & \multicolumn{3}{c|}{Kvasir} & \multicolumn{3}{c|}{EndoScene} & \multicolumn{3}{c}{ColonDB} \\ \hline
Methods & TP$^{a}$          & mDice   & mIoU    & MAE↓    & mDice    & mIoU     & MAE↓     & mDice   & mIoU    & MAE↓    \\ \hline
U-Net \cite{ronneberger2015u} & -             & 0.818   & 0.746   & 0.055   & 0.710    & 0.627    & 0.022    & 0.512   & 0.444   & 0.061   \\
ResUNet++ \cite{Jha_Smedsrud_Riegler_Johansen_Lange_Halvorsen_D} & -         & 0.821   & 0.743   & 0.048   & 0.707    & 0.624    & 0.018    & 0.483   & 0.410   & 0.064   \\ \hline
\hline
SAM-5P & -            & 0.750   & 0.645   & 0.104   & 0.656    & 0.582    & 0.139    & 0.569   & 0.482   & 0.215   \\
SAM-16P & -           & 0.719   & 0.620   & 0.140   & 0.692    & 0.613    & 0.118    & 0.548   & 0.467   & 0.228   \\ \hline
\textbf{APM-RN-5P} & 21.7M    & 0.741   & 0.645   & 0.060   & 0.781    & 0.689    & 0.016    & 0.594   & 0.501   & 0.056   \\
\textbf{APM-RN-16P} & 21.7M   & \textbf{0.797}   & \textbf{0.706}   & \textbf{0.046}   & 0.789    & 0.694    & \textbf{0.013}    & 0.595   & 0.502   & \textbf{0.051}   \\ \hline
\textbf{APM-Cross-5P}  & 44.3M &      0.749 &     0.648  &      0.074 &      0.711 &     0.619 &     0.059&     0.511 &    0.433  &    0.165  \\
\textbf{APM-Cross-16P} & 44.3M & 0.789   & 0.697   & 0.051   & \textbf{0.794}    & \textbf{0.699}    & 0.018    & \textbf{0.616}   & \textbf{0.517}   & 0.059   \\ \hline
\end{tabular}
}
\end{table}


\subsection{Pattern Shifting} \label{subsection: PS}

In Tab. \ref{tab: IPS1} and \ref{tab: IPS2}, we train the proposed IPS method (PaE with 4 IPS tokens) using GT-based prompt points. We perform tests using the same GT-based point prompts for SAM-related models and compare our method (SAM with IPS) to SOTA methods, SAM and SAM-Med2D.  Following the mainstream validation frameworks in the field of polyp segmentation \cite{fan2020pranet,wang2022stepwise}. the Kvasir and EndoScene benchmarks are used to validate the transfer performance of the proposed method, while the ColonDB and ETIS benchmarks are used to validate the generalization ability of the proposed method due to the domain gap with the training set. 

\begin{table}[!h]
\centering
\caption{Quantitative comparison of our IPS, SOTA methods, and other SAM-based methods. Best-in-class results are bolded, while second-best results are underlined. The 3P|5P and 3P|16P represent training our IPS on 3 GT-based prompt points and testing them with 5 and 16 ones. $^{a}$ Number of Trainable Parameters. $^{b}$ Number of Endoscopy/Dermoscopy images is utilized in the corresponding training set. $^{*}$ Some test set images from the corresponding dataset may be seen in the training process.}
\label{tab: IPS1}
\resizebox{0.9\columnwidth}{!}{
\begin{tabular}{lc|c|crrrrrrrrrr}
\hline
\multicolumn{3}{c|}{Benchmarks} &
  \multicolumn{2}{c|}{Kvasir} &
  \multicolumn{2}{c|}{EndoScene} &
  \multicolumn{2}{c}{ISIC2018} \\ \hline
\multicolumn{1}{l|}{Methods} &
\multicolumn{1}{c|}{TP$^{a}$} &
\multicolumn{1}{c|}{Images$^{b}$} &
  mDice &
  \multicolumn{1}{c|}{mIoU} &
  mDice &
  \multicolumn{1}{c|}{mIoU} &
  mDice &
  mIoU \\ \hline
\multicolumn{9}{c}{SOTA Methods} \\ \hline
\multicolumn{1}{l|}{U-net \cite{ronneberger2015u}} & - & 1450/2594 &
  0.818 &
  \multicolumn{1}{c|}{0.746} &
  0.710 &
  \multicolumn{1}{c|}{0.627} &
  0.855 &
  0.785 \\
\multicolumn{1}{l|}{PraNet \cite{fan2020pranet}} & - & 1450/2594 &
  {0.898} &
  \multicolumn{1}{c|}{0.840} &
  0.835 &
  \multicolumn{1}{c|}{{\ul 0.797}} &
  0.875 &
  0.787 \\
\multicolumn{1}{l|}{TransUNet \cite{chen2021transunet}} & - & 1450/2594 &
  {\ul 0.913} &
  \multicolumn{1}{c|}{{\ul 0.857}} &
  \textbf{0.893} &
  \multicolumn{1}{c|}{0.660} &
  {\ul 0.880} &
  {\ul 0.809} \\
\multicolumn{1}{l|}{SSFormer \cite{wang2022stepwise}} & - & 1450/2594 &
  \textbf{0.926} &
  \multicolumn{1}{c|}{\textbf{0.874}} &
  {\ul 0.887} &
  \multicolumn{1}{c|}{\textbf{0.821}} &
  \textbf{0.919} &
  \textbf{0.861} \\ \hline
\multicolumn{9}{c}{SAM-based Methods} \\ \hline
\multicolumn{1}{l|}{SAM-3P \cite{kirillov2023segment}} & - & - &
  0.589 &
  \multicolumn{1}{c|}{0.471} &
  0.513 &
  \multicolumn{1}{c|}{0.414} &
  0.489 &
  0.367 \\
\multicolumn{1}{l|}{SAM-5P} & - & - &
  0.750 &
  \multicolumn{1}{c|}{0.645} &
  0.656 &
  \multicolumn{1}{c|}{0.582} &
  0.687 &
  0.569 \\
\multicolumn{1}{l|}{SAM-16P} & - & - &
  0.719 &
  \multicolumn{1}{c|}{0.620} &
  0.692 &
  \multicolumn{1}{c|}{0.613} &
  0.738 &
  0.624 \\
\multicolumn{1}{l|}{Med2D-3P \cite{cheng2023sam}} &  184.5M & 5838/7935 &
  $^*$0.821 &
  \multicolumn{1}{c|}{$^*$0.735} &
  0.697 &
  \multicolumn{1}{c|}{0.597} &
  $^*$0.872 &
  $^*$0.803 \\
\multicolumn{1}{l|}{Med2D-5P} & 184.5M & 5838/7935 &
  $^*$0.822 &
  \multicolumn{1}{c|}{$^*$0.735} &
  0.722 &
  \multicolumn{1}{c|}{0.623} &
  $^*$0.884 &
  $^*$0.813 \\
\multicolumn{1}{l|}{Med2D-16P} &  184.5M & 5838/7935 &
  $^*$0.832 &
  \multicolumn{1}{c|}{$^*$0.748} &
  0.727 &
  \multicolumn{1}{c|}{0.620} &
  {\ul $^*$0.893} &
  {\ul $^*$0.823} \\ \hline
\multicolumn{1}{l|}{\textbf{IPS-3P}} & 1.3M & 1450/2594 &
  0.797 &
  \multicolumn{1}{c|}{0.704} &
  0.806 &
  \multicolumn{1}{c|}{0.718} &
  0.854 &   
  0.762 \\
\multicolumn{1}{l|}{\textbf{IPS-3P|5P}} & 1.3M & 1450/2594 &
  0.821 &
  \multicolumn{1}{c|}{0.732} &
  0.804 &
  \multicolumn{1}{c|}{0.716} &
  0.865 &
  0.774 \\
\multicolumn{1}{l|}{\textbf{IPS-3P|16P}} & 1.3M & 1450/2594 &
  0.843 &
  \multicolumn{1}{c|}{0.752} &
  0.815 &
  \multicolumn{1}{c|}{0.721} &
  0.874 &
  0.785 \\
\multicolumn{1}{l|}{\textbf{IPS-5P}} & 1.3M & 1450/2594 &
  {\ul 0.855} &
  \multicolumn{1}{c|}{{\ul 0.772}} &
  {\ul 0.854} &
  \multicolumn{1}{c|}{{\ul 0.764}} &
  {0.889} &
  {0.808} \\
\multicolumn{1}{l|}{\textbf{IPS-16P}} & 1.3M & 1450/2594 &
  \textbf{0.902} &
  \multicolumn{1}{c|}{\textbf{0.835}} &
  \textbf{0.888} &
  \multicolumn{1}{c|}{\textbf{0.810}} &
  \textbf{0.915} &
  \textbf{0.847} \\ \hline
\end{tabular}
}
\end{table}

As shown in Tab. \ref{tab: IPS1}, our proposed IPS significantly improves the performance of SAM on familiar benchmarks, achieving mDice improvements of 10\%-21\% in Kvasir, 19\%-29\% in EndoScene, and 18\%-37\% in ISIC2018.  Compared to SAM-Med2D, which is trained on a large-scale medical dataset and utilizes a complex training strategy, our method also achieves a remarkable improvement. 

Experimental results in Tab. \ref{tab: IPS2} indicate that the IPS method adequately leverages SAM's powerful generalization and feature expression capabilities. Even in the unfamiliar and challenging benchmarks of ColonDB and ETIS, IPS enables SAM to achieve SOTA performance. Using SSFormer as a baseline, we obtain up to 10\% and 9\% improvement in ColonDB and ETIS, respectively. Moreover, compared to vanilla SAM, IPS can achieving mDice improvements of 25\%-33\% in ColonDB, 20\%-33\% in ETIS. It is promising that the above performance improvement depends only on a tiny number of trainable parameters (1.3M). Those demonstrate that the IPS can efficiently transfer SAM to the target task and achieve SOTA-level performance with powerful generalization ability. This makes the low-cost rapid deployment of SAM promising.

\begin{table}[!h]
\centering
\caption{Generalization performance comparison of our IPS, SOTA methods, and other SAM-based methods. Best-in-class results are bolded, while second-best results are underlined.}
\label{tab: IPS2}
\resizebox{0.7\columnwidth}{!}{
\begin{tabular}{lc|c|crrrrrrrrrr}
\hline
\multicolumn{3}{c|}{Benchmarks} &
  \multicolumn{2}{c|}{ColonDB} &
  \multicolumn{2}{c}{ETIS} \\ \hline
\multicolumn{1}{l|}{Methods} &
\multicolumn{1}{c|}{TP$^{a}$} &
\multicolumn{1}{c|}{Images$^{b}$} &
  mDice &
  \multicolumn{1}{c|}{mIoU} &
  mDice &
  mIoU \\ \hline
\multicolumn{7}{c}{SOTA Methods} \\ \hline
\multicolumn{1}{l|}{U-net \cite{ronneberger2015u}} & - & 1450/2594 &
  0.512 &
  \multicolumn{1}{c|}{0.444} &
  0.398 &
  \multicolumn{1}{c}{0.335} \\
\multicolumn{1}{l|}{PraNet \cite{fan2020pranet}} & - & 1450/2594 &
  0.712 &
  \multicolumn{1}{c|}{0.640} &
  0.628 &
  \multicolumn{1}{c}{0.567} \\
\multicolumn{1}{l|}{TransUNet \cite{chen2021transunet}} & - & 1450/2594 &
  \textbf{0.781} &
  \multicolumn{1}{c|}{\textbf{0.699}} &
  {\ul 0.731} &
  \multicolumn{1}{c}{{\ul 0.624}} \\
\multicolumn{1}{l|}{SSFormer \cite{wang2022stepwise}} & - & 1450/2594 &
  {\ul 0.772} &
  \multicolumn{1}{c|}{{\ul 0.697}} &
  \textbf{0.767} &
  \multicolumn{1}{c}{\textbf{0.698}} \\ \hline
\multicolumn{7}{c}{SAM-based Methods} \\ \hline
\multicolumn{1}{l|}{SAM-3P \cite{kirillov2023segment}} & - & - &
  0.447 &
  \multicolumn{1}{c|}{0.356} &
  0.464 &
  \multicolumn{1}{c}{0.381} \\
\multicolumn{1}{l|}{SAM-5P} & - & - &
  0.569 &
  \multicolumn{1}{c|}{0.482} &
  0.541 &
  \multicolumn{1}{c}{0.472} \\
\multicolumn{1}{l|}{SAM-16P} & - & - &
  0.548 &
  \multicolumn{1}{c|}{0.467} &
  0.524 &
  \multicolumn{1}{c}{0.455} \\
\multicolumn{1}{l|}{Med2D-3P \cite{cheng2023sam}} &  184.5M & 5838/7935 &
  0.689 &
  \multicolumn{1}{c|}{0.588} &
  0.633 &
  \multicolumn{1}{c}{0.524} \\
\multicolumn{1}{l|}{Med2D-5P} & 184.5M & 5838/7935 &
  0.686 &
  \multicolumn{1}{c|}{0.576} &
  0.677 &
  \multicolumn{1}{c}{0.571} \\
\multicolumn{1}{l|}{Med2D-16P} &  184.5M & 5838/7935 &
  0.685 &
  \multicolumn{1}{c|}{0.575} &
  0.622 &
  \multicolumn{1}{c}{0.514} \\ \hline
\multicolumn{1}{l|}{\textbf{IPS-3P}} & 1.3M & 1450/2594 &
  0.724 &
  \multicolumn{1}{c|}{0.618} &
  0.659 &
  \multicolumn{1}{c}{0.569} \\
\multicolumn{1}{l|}{\textbf{IPS-3P|5P}} & 1.3M & 1450/2594 &
  0.761 &
  \multicolumn{1}{c|}{0.655} &
  0.727 &
  \multicolumn{1}{c}{0.626} \\
\multicolumn{1}{l|}{\textbf{IPS-3P|16P}} & 1.3M & 1450/2594 &
  0.783 &
  \multicolumn{1}{c|}{0.678} &
  0.763 &
  \multicolumn{1}{c}{0.674} \\
\multicolumn{1}{l|}{\textbf{IPS-5P}} & 1.3M & 1450/2594 &
  {\ul 0.819} &
  \multicolumn{1}{c|}{{\ul 0.716}} &
  {\ul 0.801} &
  \multicolumn{1}{c}{{\ul 0.704}} \\
\multicolumn{1}{l|}{\textbf{IPS-16P}} & 1.3M & 1450/2594 &
  \textbf{0.874} &
  \multicolumn{1}{c|}{\textbf{0.789}} &
  \textbf{0.854} &
  \multicolumn{1}{c}{\textbf{0.770}} \\ \hline
\end{tabular}
}
\end{table}

As demonstrated in Fig. \ref{fig:IPS-points} and Tab. \ref{tab: IPS1} and \ref{tab: IPS2}, SAM and SAM-Med2D struggle to handle exposures and insignificant small-volume targets well, even when provided with more fine-grained guidance through an increased number of points (\textit{i.e.} more prompt points). In contrast, our proposed IPS significantly optimizes its predicted masks through non-invasive pattern shifting based on the same prompts (Fig. \ref{fig:IPS-points}). Furthermore, the results in Tab. \ref{tab: IPS1}  and \ref{tab: IPS2} demonstrate that even if the number of prompt points used in training does not match the number of points used in testing, the IPS can still handle these information gaps well.

\subsection{ProMISe framework}
\label{subsection:Promise}

To avoid random sampling of prompts and thus achieve end-to-end pattern shifting, we propose the ProMISe framework, which couples the APM and IPS for training with adaptive prompts. When tested using GT-based prompts, the end-to-end pattern shifting of ProMISe with both APMs (Cross and RN) significantly improves SAM’s performance to a practical and competitive level in MIS (Tab. \ref{tab: Sam-Rider1} and \ref{tab: Sam-Rider2}). Importantly, ProMISe maintains its interpretability using both adaptive/GT-based Euclidean prompts and keeps all of SAM’s parameters frozen, resulting in a more practical and applicable approach for real clinical scenarios.

\begin{table}[!h]
\centering
\caption{Performance of SAM with ProMISe (Cross and RN) is tested using GT-base prompts. $^{a}$ Number of Trainable Parameters. $^{b}$ Number of Endoscopy/Dermoscopy images is utilized in the corresponding training set. $^{*}$ Some test set images from the corresponding dataset may be seen in the training process.}
\label{tab: Sam-Rider1}
\resizebox{0.88\columnwidth}{!}{
\begin{tabular}{l|c|c|rr|rr|rr}
\hline
\multicolumn{3}{c|}{Benchmarks} & \multicolumn{2}{c|}{Kvasir} & \multicolumn{2}{c|}{EndoScene} & \multicolumn{2}{c}{ISIC2018} \\ \hline
Methods &  TP$^{a}$  & Images$^{b}$  & mDice & mIoU  & mDice & mIoU  & mDice & mIoU  \\ \hline
U-Net & -  & 1450/2594          & 0.818 & 0.746 & 0.710 & 0.627 & 0.855 & 0.785 \\
ResUNet++ & -  & 1450/2594       & 0.821 & 0.743 & 0.707 & 0.624 & 0.809 & 0.729 \\ \hline
\hline
SAM-5P & -  & -            & 0.750 & 0.645 & 0.656 & 0.582 & 0.687 & 0.569 \\
SAM-16P & - & -          & 0.719 & 0.620 & 0.692 & 0.613 & 0.738 & 0.624 \\ \hline
Med2D-5P & 184.5M & 5838/7935         & $^*$0.822 & $^*$0.735 & 0.722 & 0.623 & $^*$0.884 & $^*$0.813 \\
Med2D-16P & 184.5M  & 5838/7935       & $^*$0.832 & $^*$0.748 & 0.727 & 0.620 & $^*$0.893 & $^*$0.823 \\ \hline
\hline
\textbf{Cross-5P} & 45.6M  & 1450/2594  & 0.834 & 0.744 & 0.788 & 0.687 & 0.742 & 0.631 \\
\textbf{Cross-16P} & 45.6M  & 1450/2594 &  0.858 & 0.777 & 0.768 & 0.661 & 0.819 & 0.705 \\ \hline
\textbf{RN-5P} & 23.0M  & 1450/2594  & 0.776 & 0.673 & 0.705 & 0.587 & 0.798 & 0.688 \\
\textbf{RN-16P} & 23.0M  & 1450/2594  & 0.846 & 0.759 & 0.803 & 0.735 & 0.878 & 0.791 \\ \hline
\end{tabular}
}
\end{table}

\begin{table}[]
\centering
\caption{Generalization performance of SAM with ProMISe (Cross and RN) is tested using GT-base prompts.}
\label{tab: Sam-Rider2}
\resizebox{0.7\columnwidth}{!}{
\begin{tabular}{l|c|c|rr|rr}
\hline
\multicolumn{3}{c|}{Benchmarks} & \multicolumn{2}{c|}{ColonDB} & \multicolumn{2}{c}{ETIS} \\ \hline
Methods &  TP$^{a}$  & Images$^{b}$ & mDice & mIoU  & mDice & mIoU  \\ \hline
U-Net & -  & 1450/2594          & 0.512 & 0.444 & 0.398 & 0.335 \\
ResUNet++ & -  & 1450/2594       & 0.483 & 0.410 & 0.401 & 0.344 \\ \hline
\hline
SAM-5P & -  & -            & 0.569 & 0.482 & 0.541 & 0.472 \\
SAM-16P & - & -          & 0.548 & 0.467 & 0.524 & 0.455 \\ \hline
Med2D-5P & 184.5M & 5838/7935         & 0.686 & 0.576 & 0.677 & 0.571 \\
Med2D-16P & 184.5M  & 5838/7935       & 0.685 & 0.575 & 0.622 & 0.514 \\ \hline
\hline
\textbf{Cross-5P} & 45.6M  & 1450/2594  & 0.744 & 0.636 & 0.678 & 0.578 \\
\textbf{Cross-16P} & 45.6M  & 1450/2594 & 0.732 & 0.626 & 0.691 & 0.591 \\ \hline
\textbf{RN-5P} & 23.0M  & 1450/2594  & 0.664 & 0.547 & 0.605 & 0.508 \\
\textbf{RN-16P} & 23.0M  & 1450/2594  & 0.735 & 0.624 & 0.673 & 0.566 \\ \hline
\end{tabular}
}
\end{table}

\subsection{Multi-Modality Experiments}

\begin{figure}[!h]
    \centering
    \includegraphics[width=\textwidth]{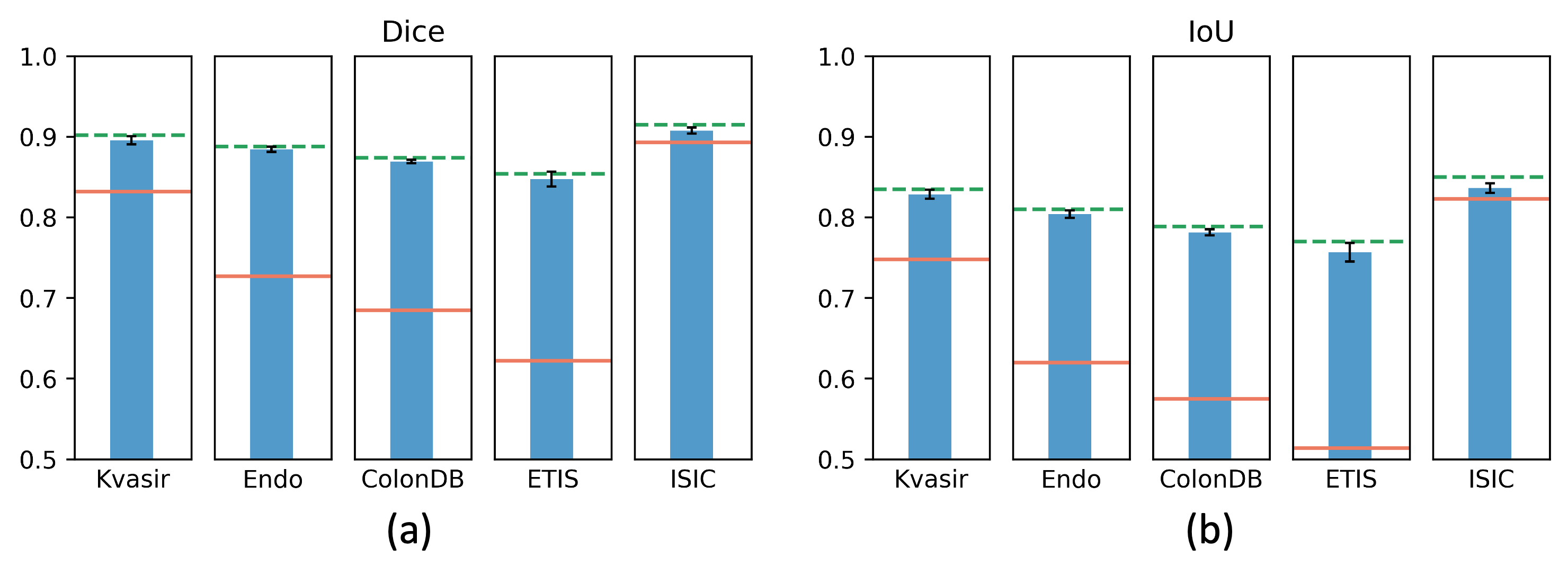}
    \caption{Multi-modality training performance of pattern shifting with 16 points (mDice scores). The corresponding results of MedSAM-2D and our proposed IPS are represented by red lines and green dashed lines, respectively, as shown in Table \ref{tab: IPS1} and \ref{tab: IPS2}. The blue bars represent the performance of IPS training with both endoscopy and dermoscopy images.}
    \label{fig:IPS-MM}
\end{figure}

To evaluate the multi-modality training potential of our proposed method, we input both endoscopy and dermoscopy together as the training set. Additionally, to verify the stability of our method in multi-modality training, we applied five different random seeds to provide GT-based prompt points during the evaluation. As shown in Fig. \ref{fig:IPS-MM}, the performance of multi-modality training differs only slightly from the results of the corresponding single-modality training (green dashed lines) and is significantly higher than MedSAM-2D (red lines). These experimental results indicate that our IPS method has the potential not only for rapid and low-cost pattern shifting to a single medical domain but also for multiple specific modalities.

Furthermore, the error bars (standard deviation) indicate that our method has excellent robustness on both Dice and IoU metrics, as the GT-based prompt points generated from five different random seeds produce relatively consistent results. This is particularly meaningful in clinical scenarios, as different clinicians may provide preferred prompt points. Using the method proposed in this paper, they can receive similar results. 

\subsection{Ablation Study}

Apart from the differences in effects brought by the individual and combined application of IPS and APM discussed in Sect. \ref{subsection:APM} - \ref{subsection:Promise}, we also find that modifications to mask tokens can significantly enhance the pattern-shifting ability of the SAM-based model. As demonstrated in Tab. \ref{tab: ablation}, adding IPS tokens to mask tokens significantly improve the performance, and the PaE module is indispensable for the IPS method to achieve SOTA results. Using IPS tokens alone represents initialization of the tokens rather than being generated from the PaE. Thus, removing the PaE provides an extremely lightweight option.

\begin{table}[!t]
\centering
\caption{Ablation study for pattern shifting with 16 points (mDice scores). Mask tokens refer to unfreezing the mask tokens in SAM's mask decoder. The optimal setting is in bold. $^{a}$ Number of Trainable Parameters.}
\label{tab: ablation}
\resizebox{\columnwidth}{!}{
\begin{tabular}{c|c|c|c|c|c|c|c|c}
\hline
Mask tokens & IPS tokens & PaE & Frozen SAM & TP$^{a}$ & Kvasir & EndoScene & ColonDB & ETIS  \\ \hline
\Checkmark & & & \XSolidBrush & 1.02K & 0.581&       0.796 &     0.756 &   0.726 \\
 & \Checkmark & & \Checkmark & 1.02K & 0.862  & 0.817     & 0.802   & 0.778 \\
 & \Checkmark & \Checkmark & \Checkmark & 1.29M & \textbf{0.902} & \textbf{0.888} & \textbf{0.874} & \textbf{0.854} \\ \hline
\end{tabular}
}
\end{table}
 
\section{Conclusions}
In this paper, we propose a novel adaptive prompt generation module, Auto-Prompting Module (APM), which improves the transfer-free performance of SAM in the target domain by generating optimal Euclidean prompts. In addition, we propose Incremental Pattern Shifting (IPS), which enables non-fine-tuned pattern shifting to improve SAM’s performance in unfamiliar domains, achieving SOTA and competitive results. Furthermore,  we couple IPS with APM to propose the ProMISe framework, which can realizes end-to-end pattern shifting to improve training efficiency and stability. We conducted experiments in endoscopic and dermoscopic benchmark datasets to demonstrate the usefulness and promise of our proposed methods. Benefiting from IPS, increasing prompt point number results in significant performance gains for SAM, which may be further extended to include scrawl, sketch or coarse prompts. More importantly, our results prove that a fine-tuning-based approach is not necessarily optimal for utilizing a foundation model like SAM.

\section{Limitations and future works}

Although IPS has achieved promising performance in medical image segmentation tasks in endoscopic and dermoscopic modalities, the medical modalities involved in this paper still need to be increased to validate the effect of IPS in the entire medical image segmentation domain. Moreover, while our proposed APM approach can effectively improve the performance of transfer-free SAM in medical image segmentation, it performs sub-optimally when coupled with IPS. This suggests that we must trade between the end-to-end framework and performance.

 Based on the above limitations, we plan to extend the ProMISe approach to as many medical image modalities ( including 2D and 3D) as possible. In order to improve the reliability of ProMISe, we plan to optimize the APM method while attempting a lightweight, non-invasive shifting of the representation encoder. Moreover, to improve the extensibility and practicality of ProMISe and IPS, we plan to enrich further the prompt types, such as bounding boxes, scribble, sketch, and text.

\bibliographystyle{abbrvnat}
{\small
\bibliography{NeurIPS2024}
}



\end{document}